\documentclass[sigplan,nonacm]{acmart}
\usepackage{booktabs}
\usepackage{microtype}
\usepackage{threeparttable}
\usepackage{makecell}
\usepackage{array}
\usepackage{placeins}
\usepackage{enumitem}
\renewcommand{\paragraph}[1]{\noindent\textbf{#1}\ }
\title{OmniTide: Co-Designing Algorithms and Systems for Efficient On-Device Omni-LLM Streaming}

\newcommand{\pkuaffiliation}{%
  \affiliation{%
    \institution{Key Lab of High Confidence Software Technologies (Peking University)}%
    \city{Beijing}%
    \country{China}}}
\newcommand{\buptaffiliation}{%
  \affiliation{%
    \institution{State Key Laboratory of Networking and Switching Technology (BUPT)}%
    \city{Beijing}%
    \country{China}}}
\author{Zongshang Shen}
\pkuaffiliation
\email{shenzongshang26@stu.pku.edu.cn}
\author{Wangsong Yin}
\pkuaffiliation
\author{Daliang Xu}
\buptaffiliation
\email{xudaliang@bupt.edu.cn}
\author{Mengwei Xu}
\buptaffiliation
\author{Xuanzhe Liu}
\pkuaffiliation
\renewcommand{\shortauthors}{Shen et al.}

\makeatletter
\renewcommand{\@mkauthors@iii}{%
  \global\setbox\mktitle@bx=\vbox{%
  \unvbox\mktitle@bx\par\medskip
  \centering
  \normalfont\fontsize{12}{15}\selectfont
  Zongshang Shen\textsuperscript{\(\blacklozenge\)},
  Wangsong Yin\textsuperscript{\(\blacklozenge\)},
  Daliang Xu\textsuperscript{\(\lozenge\)\#},
  Mengwei Xu\textsuperscript{\(\lozenge\)},
  Xuanzhe Liu\textsuperscript{\(\blacklozenge\)\#}\par
  \vspace{3pt}
  \fontsize{10}{12}\selectfont
  \textsuperscript{\(\blacklozenge\)}Key Lab of High Confidence Software Technologies (Peking University), Beijing, China\par
  \textsuperscript{\(\lozenge\)}State Key Laboratory of Networking and Switching Technology (BUPT), Beijing, China\par
  \vspace{2pt}
  \href{mailto:shenzongshang26@stu.pku.edu.cn}{shenzongshang26@stu.pku.edu.cn},
  \href{mailto:xudaliang@bupt.edu.cn}{xudaliang@bupt.edu.cn}\par
  \vspace{12pt}}}
\makeatother

\begin{document}
\begin{abstract}
On-device streaming omni-modal inference safeguards user privacy and eliminates prohibitive per-token API costs, but faces a critical bottleneck: the continuous influx of multimodal data rapidly exhausts constrained memory and compute budgets via monotonic KV cache growth. Existing sparse attention methods fall short, either incurring prohibitive online estimation latency or destroying interleaved cross-modal context, while failing to resolve physical memory fragmentation. We present \texttt{OmniTide}, the first algorithm-system co-design tailored for efficient on-device streaming omni-modal inference. Driven by the observation of \textit{modality-aware structural sparsity}, \texttt{OmniTide} adopts a \textit{unit}-based abstraction with two components: (1) At the algorithm level, \texttt{OmniPick} logically retains critical multimodal context based on unit boundaries and modality importance to preserve task accuracy; (2) At the system level, \texttt{OmniPage} physically partitions the cache by retention likelihood and dynamically compacts surviving sparse tokens, minimizing both memory fragmentation and data-movement overhead. Extensive evaluations across three streaming benchmarks and two consumer-device architectures show that \texttt{OmniTide} achieves up to $12.72\times$ kernel speedups and $2.40\times$ lower stream-loop latency. On StreamingBench, it improves accuracy by up to 18.0 percentage points over sliding-window baselines at comparable session cost. OmniPage further reduces the physical KV span by up to 26.7\% relative to native logical eviction, unlocking real-time, infinite-context streaming on edge devices.
\end{abstract}

\maketitle
\raggedbottom

\section{Introduction}
Omni-modal models (or \emph{omni models}) unify cross-modal comprehension and generation---spanning text, speech, images, and video---within an end-to-end architecture~\cite{xu2025qwen25omni}. Today, these models power real-time interactive agents, such as \emph{ChatGPT Voice \& Vision}~\cite{openai2024voicevideo}, \emph{Google Gemini Live}~\cite{sheth2025geminilive}, and \emph{Doubao Real-time Assistant}~\cite{bytedance2026seedrealtime}. Unlike traditional single-turn systems, omni models engage in continuous, full-duplex interactions. For instance, a user can stream live camera feeds while speaking naturally; concurrently, the model perceives the audiovisual stream, delivers real-time spoken responses, and dynamically adapts to follow-up corrections.

\paragraph{Bottleneck of on-device streaming omni inference.} Deploying omni models directly on consumer devices safeguards privacy and avoids network or per-token costs~\cite{cui2026minicpmo45,zhang2026vlmcache}. However, \textit{on-device streaming inference faces a severe bottleneck: Continuously arriving multimodal data rapidly exhausts constrained memory and compute budgets.} Because queries may reference any past observation, the runtime incrementally appends temporal chunks (e.g., interleaved video and audio) to the session's key--value (KV) cache, causing monotonic growth. Our preliminary experiments with MiniCPM-o-4.5~\cite{cui2026minicpmo45} on an Apple M2 Pro highlight this severity. High-resolution frames ($640$--$2,304$ tokens/image) and continuous audio ($10$--$25$ tokens/s) inject hundreds of tokens per second. Sustaining just 6.25 minutes of input inflates the context to nearly 80K tokens, consuming nearly 23~GiB of memory---far exceeding the 16~GiB capacity typical of consumer devices. Simultaneously, per-chunk prefill latency degrades by $2.5\times$ (28~s to 69~s), driven primarily by a $19.5\times$ surge in attention-kernel compute time. Therefore, bounding KV history growth while preserving critical multimodal context is a first-order systems challenge for sustained on-device omni inference.

\paragraph{Sparse attention opportunity and gaps in streaming omni inference.} 
Recent methods exploit the inherent sparsity of attention, retaining only a critical subset of tokens to bound KV cache growth without quality degradation~\cite{zhang2023h2o,xiao2024streamingllm,jiang2024minference}. However, existing approaches fall short for on-device omni-modal streams. First, \emph{attention-score-based} selection (e.g., H$_2$O, Quest)~\cite{jiang2024minference,lai2025flexprefill,tang2024quest,zhang2023h2o,liu2023scissorhands} dynamically identifies important tokens by computing queries against the history or cache metadata before eviction. Yet, this online estimation introduces unacceptable latency overheads (accounting for 33.02\% of attention time in our 80K-token profile). Second, \emph{position-based} retention policies (e.g., StreamingLLM)~\cite{jiang2023mistral,xiao2024streamingllm,xu2026streamingvlm} avoid online computation by statically keeping tokens based on recency or fixed sink positions. While fast, they are designed for unimodal text or vision and lack the semantic awareness needed to handle highly interleaved, redundant omni-modal streams. Applying them directly to our workloads incurs a quality drop of up to 17.1 percentage points on StreamingBench. More critically, across both approaches, theoretical sparsity does not automatically yield hardware efficiency: retaining scattered tokens causes severe memory fragmentation, and dynamic KV compaction introduces prohibitive memory-copying overheads (up to 59.6\% of stream-loop time in our profiling).

\paragraph{OmniTide: Algorithm-system co-design.} 
We present \texttt{Omni\-Tide}, the first system tailored for efficient on-device streaming omni-modal inference. Its goal is to build a training-free, modality-aware bounded-history system that preserves the generation quality of full-history attention while translating logical retention into actual reductions in KV-memory footprint and attention work. To achieve this, OmniTide must answer two critical questions: 
(i) \textit{Accuracy (token selection):} How can we identify the minimal subset of critical tokens from a highly interleaved, redundant cross-modal stream without costly online computation? 
(ii) \textit{Efficiency (KV management):} How can we physically organize the surviving KV cache to prevent severe memory fragmentation and avoid prohibitive dynamic compaction overheads?

To address these challenges, our core insight is that token selection and KV cache management must be co-designed around the abstraction of a \textit{unit}---the natural multimodal context boundary corresponding to the interleaved image, audio, and text tokens within a single temporal chunk. Based on this abstraction, OmniTide protects the system prefix, retains recent units in full, and selectively retains spans from older history. It introduces two novel techniques:

\paragraph{OmniPick token selection algorithm.} OmniPick is driven by three key observations of structured attention in omni-modal streams (detailed in \S\ref{subsec:structured-attention}). First, \textit{unit- and modality-local sinks} dictate that structural and boundary tokens must be preserved. Second, \textit{recent-context concentration} motivates retaining a sliding window of complete, recent units. Third, \textit{modality-asymmetric attention} reveals that certain modalities (e.g., audio) carry higher information density and warrant higher retention priority. Putting this together, when the memory budget is exceeded, OmniPick deterministically secures essential sinks and a recent window of intact units. It then allocates the remaining budget to audio, text, and other tokens based on their modality importance. Finally, the surviving tokens are logically retained and reindexed with updated relative position encodings.

\paragraph{OmniPage KV management system.} While OmniPick establishes logical retention, OmniPage translates this into physical efficiency. Our core insight is that grouping tokens by their retention likelihood reduces future memory fragmentation. Therefore, OmniPage employs a partitioned storage strategy: it physically groups KV entries by retention state, organizing them into pages aligned with attention tiles. When the history exceeds the high watermark, OmniPick evicts unselected tokens, and OmniPage plans selective migration of the surviving KV entries to limit data-movement overhead. The plan is executed only if it reduces the number of active pages or subblocks, or shortens the physical span by at least a configured amount.

\paragraph{Implementation and evaluation.}
We have implemented \texttt{OmniTide} in \texttt{llama.cpp-omni}~\cite{llamacppomni2026}---adding approximately 9.6K code lines across the runtime, model runners, and backend extensions---and evaluate it on NVIDIA RTX 4090 (CUDA) and Apple M2 Pro GPU (Metal) to assess its effectiveness across consumer-device architectures. 
We evaluate three omni-modal models (Qwen2.5-Omni-3B/7B~\cite{xu2025qwen25omni} and MiniCPM-o-4.5~\cite{cui2026minicpmo45}) on three streaming benchmarks (StreamingBench~\cite{lin2025streamingbench}, SVBench~\cite{yang2025svbench}, and LiveSports-3K-CC~\cite{chen2025livecc}). 
To rigorously contextualize our performance, we compare \texttt{OmniTide} against five baselines spanning static retention (Full Context), position-based sliding (Unit-level FIFO, Token-level sliding, Streaming\-LLM~\cite{xiao2024streamingllm}), and score-based eviction (H$_2$O~\cite{zhang2023h2o}). Across 355 StreamingBench video sessions with a median context length of around 70K tokens, OmniTide achieves up to $12.72\times$ Flash\-Attention-kernel speed\-up and $2.40\times$ speedup in post-initiali\-zation stream-loop execution compared to the full-context baseline.
Across memory budgets, \texttt{OmniPick} improves StreamingBench accuracy by up to 18.0 points over sliding windows, while \texttt{OmniPage} reduces the physical KV span by 26.7\%. Crucially, under a 4~GiB consumer budget, \texttt{OmniTide} processes a 241-second stream (220K tokens, Qwen2.5-Omni-7B) using just 0.164~GiB of physical KV, whereas full-context retention balloons to 3.5~GiB in only 67 seconds, demonstrating \texttt{OmniTide} 's capacity to sustain vastly longer streams with a minimal active footprint.
 Ultimately, by synergizing logical token selection with physical KV management, \texttt{Omni\-Tide} unlocks real-time, infinite-context streaming inference on consumer devices, generating at 109.9 tokens/s even at the maximum 64K inference window.

\noindent \textbf{Contributions} are summarized as follows:
\begin{itemize}[leftmargin=1.25em,labelsep=0.45em,nosep]
    \item We characterize the KV-cache bottleneck in on-device streaming omni inference and identify \textit{modality-aware structural sparsity} (unit-boundary sinks, recent-context concentration, and modality asymmetry).
    \item We present \texttt{OmniTide}, an algorithm-system co-design that unites logical token retention based on unit boundaries (\texttt{OmniPick}) with dynamic physical KV compaction to eliminate memory fragmentation (\texttt{OmniPage}).
    \item Extensive evaluations across 3 omni models, 3 streaming benchmarks and 2 consumer-grade GPUs demonstrate \texttt{OmniTide}'s effectiveness in accuracy, latency, and memory footprints.
\end{itemize}

\section{Background}

\subsection{On-device streaming omni-modal inference}
\label{subsec:streaming-inference}
Real-time applications, such as meeting assistants and interactive
video assistants, process continuous audio and video streams
alongside user text~\cite{chen2024videollmonline,chen2025livecc}.
To respond to new inputs, the model may need earlier observations
and its own previous responses as context.
Running these applications on-device keeps private media local
and avoids the network latency of remote inference.

\begin{figure}[t]
  \centering
  \includegraphics[width=0.97\columnwidth]{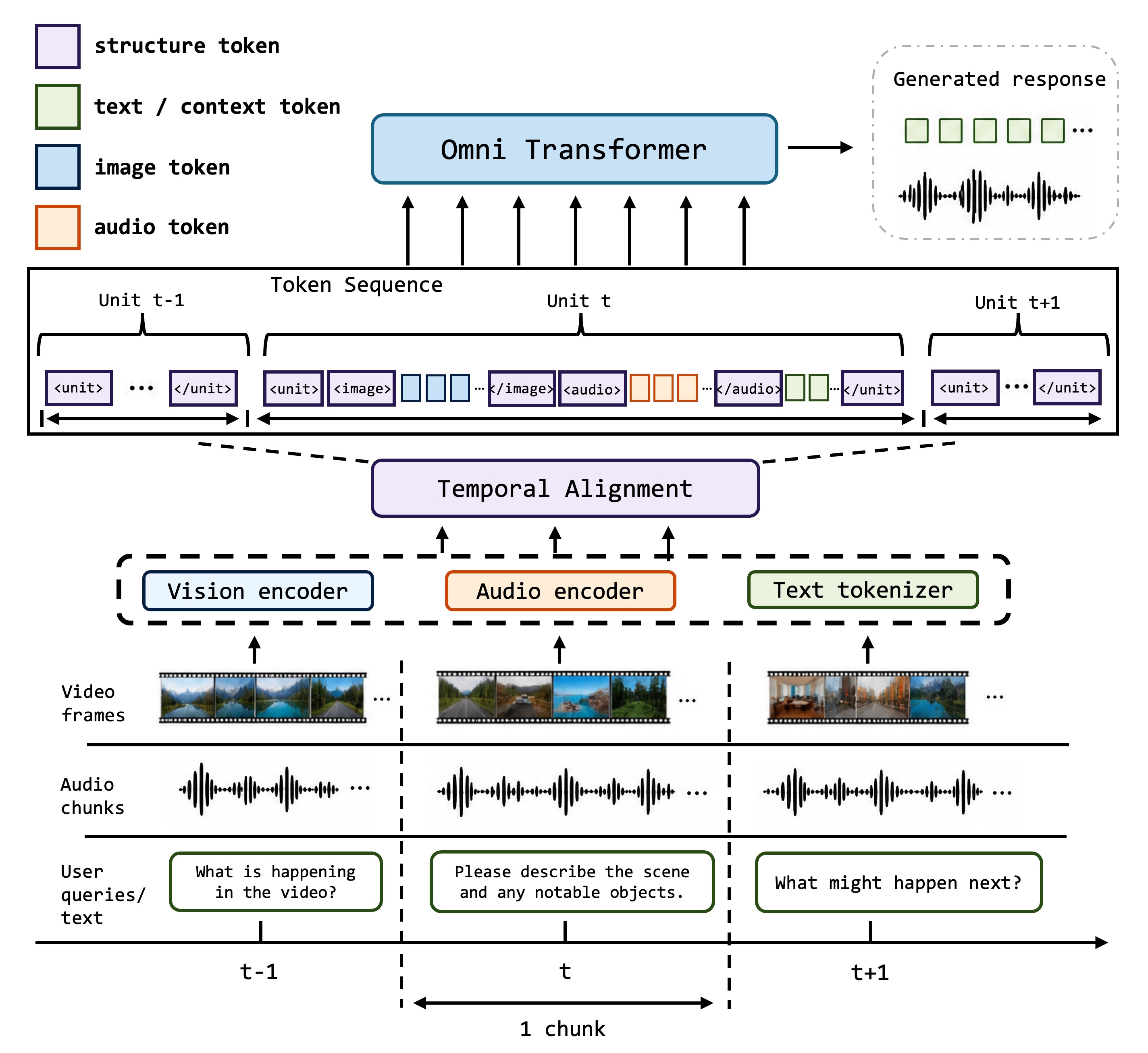}
  \caption{Streaming omni-modal inference with modality-specific encoders and an LLM backbone. Explicit unit delimiters are illustrated using MiniCPM-o-4.5 as an example.}
  \Description{A streaming omni-modal pipeline showing modality-specific encoders, chunk-sized units, and an omni transformer.}
  \label{fig:omni-input}
\end{figure}

\paragraph{Omni model architecture.}
The term \emph{omni model} denotes a unified model that processes text, images, audio, and video. A common approach is to encode non-text inputs into representations that an LLM can process~\cite{lin2024videollava,tang2024salmonn, sun2024videosalmonn,han2024onellm}. As illustrated in Figure~\ref{fig:omni-input}, a typical omni-modal model~\cite{cui2026minicpmo45,xu2025qwen25omni} combines modality-specific encoders with a large language model (LLM) backbone. The backbone processes this sequence through \emph{prefill}
and generates responses through \emph{decode}~\cite{zhong2024distserve}.
During prefill, it computes key and value (KV) states for the
new input and appends them to the session's KV cache.
During decode, it generates text one token at a time, reusing
the cached states and adding KV states as generated tokens
are processed. The cache persists across successive inputs and responses, allowing the model to attend to earlier context.
These models also support spoken responses through
a speech-generation module~\cite{xu2025qwen25omni,cui2026minicpmo45}.

\paragraph{Streaming units.}
To support incremental processing, as shown in Figure~\ref{fig:omni-input}, the unified sequence preserves the temporal organization of incoming streams by grouping related modality content into successive chunks. We define a \emph{unit} as the token sequence associated with one streaming event, including its modality content and any associated structural markers. A unit may contain an aligned audiovisual chunk, a user text segment, or a generated response. MiniCPM-o-4.5~\cite{cui2026minicpmo45} makes unit boundaries explicit through special tokens. Qwen2.5-Omni~\cite{xu2025qwen25omni} likewise organizes audiovisual inputs into time-aligned chunks through its \emph{time-interleaving method}. Thus, our unit definition captures the models' existing chunk-level organization, regardless of whether dedicated unit delimiters are present.

\subsection{On-device omni inference bottlenecks}
\label{subsec:inference-bottlenecks}
Prior studies and systems highlight memory and compute constraints in LLM inference on personal devices~\cite{song2024powerinfer,laskaridis2024melt}. Quantization reduces model-weight memory and can lower computation costs~\cite{frantar2023gptq,lin2024awq,xiao2023smoothquant}. However, fitting the model in memory is only part of the problem. With full-history retention, the KV cache grows as new inputs and responses arrive. A longer history requires more KV storage and more attention computation for each new input.
Audiovisual inputs can grow this history rapidly.
Table~\ref{tab:modality-token-counts} estimates the input rate for a stream containing one $1344\times1344$ image and one second of audio per unit, at one unit per second. Under this assumption, MiniCPM models receive roughly 700 tokens per unit, or 42K tokens per minute, while Qwen2.5-Omni-7B receives roughly 2,300 tokens per unit, or 140K tokens per minute. These estimates exclude user text and generated responses, which further increase the history.

\begin{table}[t]
  \centering
  \caption{LLM input token counts and estimated audiovisual token growth at one unit per second~\cite{openbmb2025minicpmo26,cui2026minicpmo45,xu2025qwen25omni}.}
  \label{tab:modality-token-counts}
  \begin{threeparttable}
    \footnotesize
    \setlength{\tabcolsep}{3pt}
    \begin{tabular}{@{}lrrrr@{}}
      \toprule
      Model & \makecell{Image\\$1344\times1344$} & \makecell{Audio\\tokens/s} & \makecell{Est.\\tokens/unit} & \makecell{Est.\\tokens/min} \\
      \midrule
      MiniCPM-o-2.6 & 640 & 25 & $\approx700$ & $\approx42\mathrm{K}$ \\
      MiniCPM-o-4.5 & 640 & 10 & $\approx700$ & $\approx42\mathrm{K}$ \\
      Qwen2.5-Omni-7B & 2,304 & 25 & $\approx2{,}300$ & $\approx140\mathrm{K}$ \\
      \bottomrule
    \end{tabular}
    \begin{tablenotes}[flushleft]
      \footnotesize
      \item[] MiniCPM inputs use one global view and a $3\times3$ grid of local slices.
    \end{tablenotes}
  \end{threeparttable}
\end{table}

\begin{figure}[t]
  \centering
  \includegraphics[width=\columnwidth]{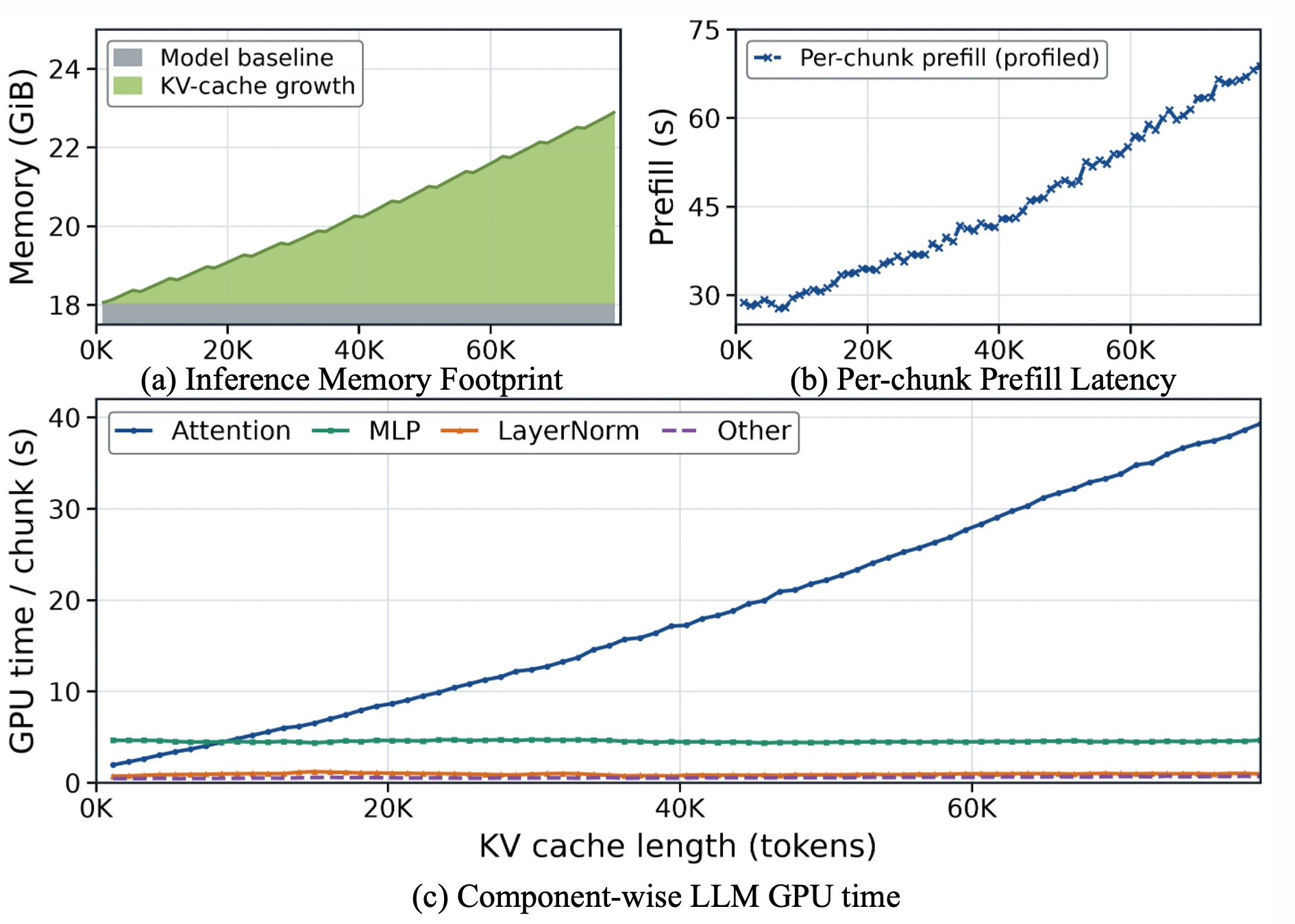}
  \caption{Memory and latency growth with full-history retention on MiniCPM-o-4.5. The model runs on Apple M2 Pro using Q4\_K\_M LLM weights and FP16 vision/audio projectors, with Metal backend and FlashAttention enabled. Around 24~GB of unified memory is available for inference.}
  \Description{Three plots showing inference memory reaching approximately 23 GiB, per-chunk prefill latency rising from 28 to 69 seconds, and attention time increasing from 2 to 39 seconds as the KV cache approaches 80,000 tokens.}
  \label{fig:motivation}
\end{figure}

To quantify these costs, we run MiniCPM-o-4.5 on an Apple M2 Pro with full-history retention, processing a 375-s LongVALE~\cite{geng2025longvale} video in 75 consecutive 5-s audiovisual chunks. By the end of the stream, the KV cache approaches 80K tokens. Despite a substantial on-device memory budget, retaining just 6.25 minutes of audiovisual history brings the total memory footprint to approximately 23~GiB, leaving little headroom for continued streaming (Figure~\ref{fig:motivation}(a)). Per-chunk prefill latency increases from 28 to 69~s ($2.5\times$; Figure~\ref{fig:motivation}(b)). Attention accounts for most of this increase: its per-chunk execution time rises from 2 to 39~s, while MLP time stays near 4.5~s and other components change little (Figure~\ref{fig:motivation}(c)). Thus, retaining the full history increases both memory use and the time needed to process each new chunk.

\section{Motivation}
\begin{figure*}[t]
  \centering
  \includegraphics[width=\textwidth]{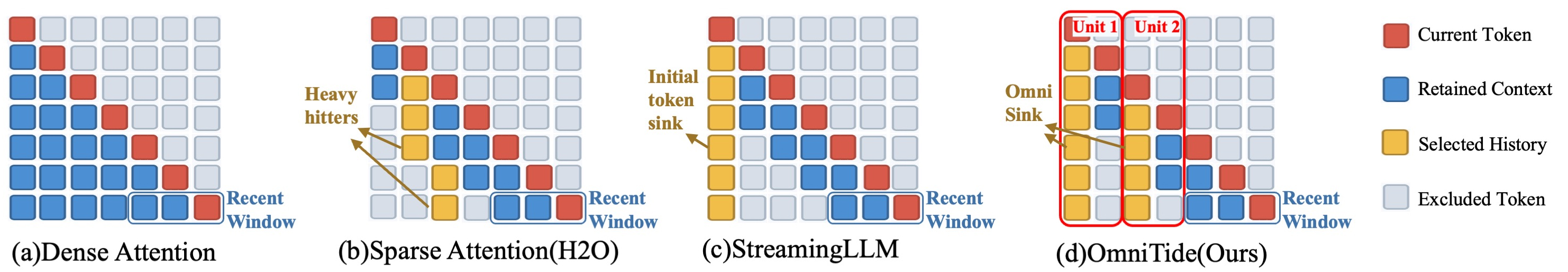}
  \caption{History-selection strategies: (a) dense attention, (b) H$_2$O heavy-hitter retention, (c) StreamingLLM's global sink prefix and recent window, and (d) OmniTide's unit-aware local sinks and recent context.}
  \Description{Four schematic attention patterns comparing dense history, H2O heavy hitters, StreamingLLM, and OmniTide with explicit unit boundaries.}
  \label{fig:history-policies}
\end{figure*}

Figure~\ref{fig:history-policies} compares various history-selection strategies proposed to address the long-context bottleneck, highlighting the inherent tension between bounding cache size and preserving multimodal context.

\paragraph{Dense attention.}
Dense attention retains the KV states of all preceding tokens, making the complete history available to subsequent queries (Figure~\ref{fig:history-policies}(a)). Although it avoids information loss from eviction, its cache size and attention cost grow with the session, leading to the on-device bottlenecks characterized in \S\ref{subsec:inference-bottlenecks}.

\paragraph{Attention-score-based selection.}
These methods use attention scores as a proxy for token importance, selecting historical KV entries under a cache budget. They differ in when scores are collected and how retention budgets are distributed. One line of work uses attention observed during decoding to guide eviction~\cite{zhang2023h2o,liu2023scissorhands,oren2024tova}; for example, H$_2$O retains recent tokens and accumulated-attention heavy hitters (Figure~\ref{fig:history-policies}(b)). Another uses prompt attention to identify head-specific context to retain~\cite{li2024snapkv,ge2024fastgen}, with SnapKV collecting scores from a trailing observation window. Complementary methods adapt cache allocation across layers or heads rather than assigning uniform budgets~\cite{cai2025pyramidkv,feng2025adakv,fu2025headkv}. These approaches allow selected historical evidence to survive beyond a recent window. However, obtaining attention statistics and selecting retained entries introduce runtime work beyond positional retention, with costs depending on the scoring schedule and implementation. Moreover, attention-based token ranking does not itself encode multimodal-unit boundaries or the association between modality content and structural spans. Thus, even high-scoring retained tokens need not preserve complete audiovisual units or their boundary context, motivating a retention policy that explicitly represents these relationships.

\paragraph{Position-based retention.}
These methods retain a recent window and optionally protect designated tokens outside it, evicting older KV entries without online attention-score estimation. Local-window designs use rolling buffers to bound cache growth~\cite{jiang2023mistral}, while prefix-augmented schemes combine initial tokens with recent history~\cite{xiao2024streamingllm,han2024lminfinite}; StreamingLLM illustrates this pattern in Figure~\ref{fig:history-policies}(c). Extensions retain separator tokens beyond the local window~\cite{chen2025sepllm} or assign different windows to different modalities~\cite{xu2026streamingvlm}. For example, StreamingVLM retains initial sinks, a longer text history, and a shorter vision window. These rules reduce both the retained KV history and subsequent attention work. However, designs for text or vision-language streams do not directly specify retention for interleaved omni-modal units: token cutoffs can split an audiovisual unit, and protecting an initial prefix does not preserve recurring local sinks in later units. Separator protection and separate modality windows likewise do not explicitly preserve the association between aligned audiovisual content and its structural spans. This limitation motivates unit-aware retention; in our evaluation, token-level sliding reduces MiniCPM-o-4.5's StreamingBench accuracy from 74.1\% to 57.0\%, a 17.1-percentage-point drop relative to full-context retention (Table~\ref{tab:quality}).

\paragraph{OmniTide design goal.}
These limitations motivate a train\-ing-free policy that uses streaming structure without attention-score collection (Figure~\ref{fig:history-policies}(d)).
OmniTide aims to bound KV-cache growth and reduce history-dependent attention cost while preserving response quality. This goal has two parts: preserving useful context within a bounded logical history, and organizing the retained KV entries so that logical eviction can translate into physical execution savings.

\section{OmniTide overview}
\label{sec:design-overview}
\begingroup
\emergencystretch=1em

\paragraph{Key insight.} 
Streaming omni-modal inputs form \textit{units} representing concurrent events within a temporal chunk:
$$
\begin{gathered}
\texttt{<unit>}\;\texttt{<image>}\;[\text{vision tokens}]\;\texttt{</image>}\\
\texttt{<audio>}\;[\text{audio tokens}]\;\texttt{</audio>}\;[\text{text tokens}]\;\texttt{</unit>}
\end{gathered}
$$
Because units encapsulate interleaved cross-modal tokens, ignoring these boundaries dismantles temporal alignment and semantic cohesion. Thus, OmniTide uses \textit{unit} as its core abstraction for token selection and KV cache management. 

\paragraph{$\bullet$ OmniPick token selection algorithm.} 
Our attention analysis (\S\ref{subsec:structured-attention}) reveals \textit{inter-unit recency} (attention concentrates on recent complete units) and \textit{intra-unit asymmetry} (varying token importance). Consequently, OmniPick organizes retention into three groups: \textit{must-keep} (e.g., system prompts/text), \textit{selected-history} (e.g., modality-specific sink), and \textit{recent-context} (evictable). It uses these groups to deterministically retain critical multimodal context. 

\paragraph{$\bullet$ OmniPage KV management system.}
OmniPage (\S\ref{sec:omnipage}) uses OmniPick's retention
decisions to guide physical KV placement. It groups entries
by retention state in tile-aligned pages, reducing mixing between newly arriving content and older units' retained spans.
When the context history exceeds a high watermark, OmniPick evicts content toward a low-watermark target, leaving room for
subsequent inputs. After eviction, OmniPage evaluates whether
selectively relocating surviving entries would reduce the
number of occupied pages or subblocks, or shorten physical
extent presented to attention. It applies the migration plan
only when the backend-specific condition is met.

\begin{figure}[t]
  \centering
  \includegraphics[width=\columnwidth]{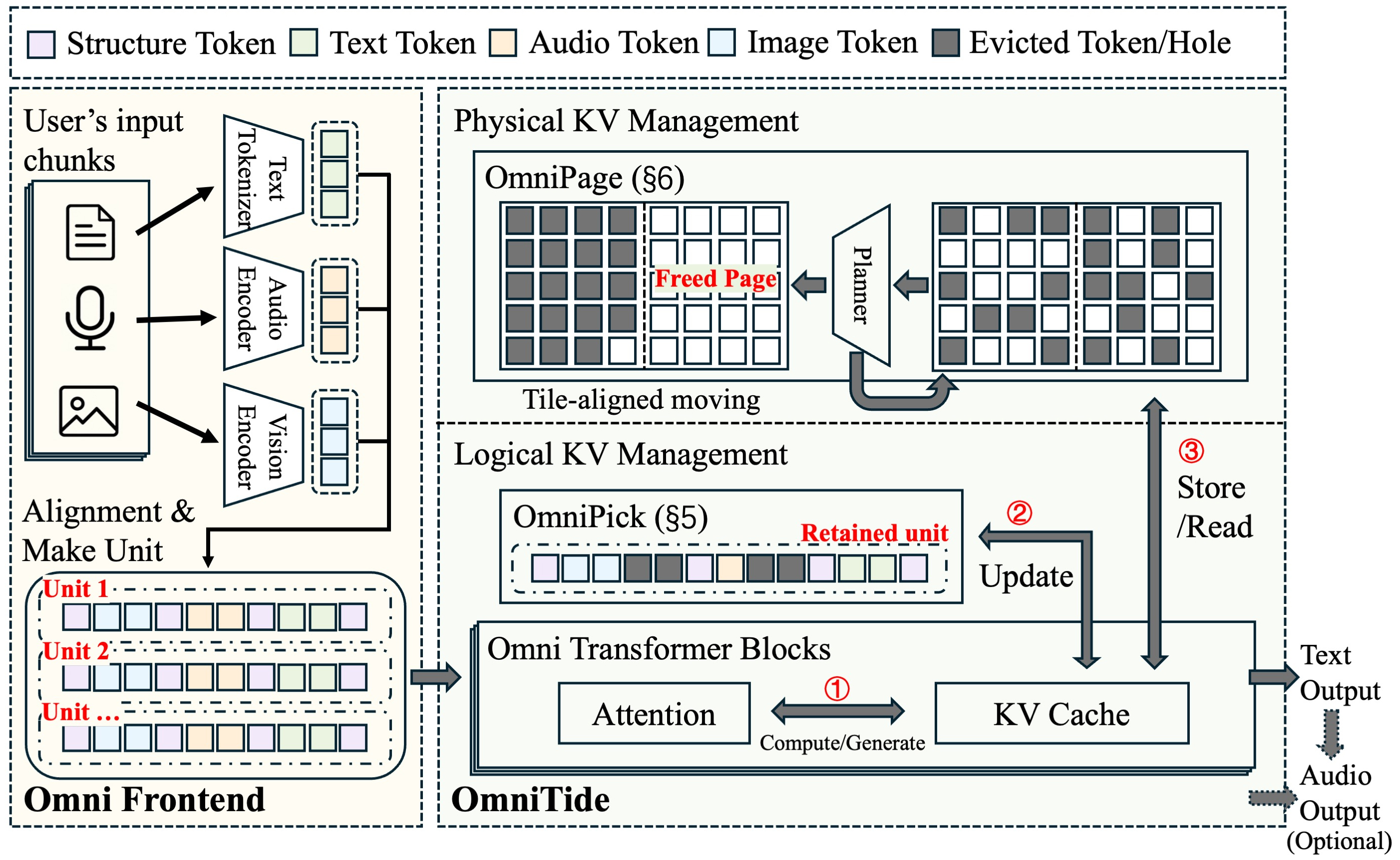}
  \caption{OmniTide runtime overview. The frontend assembles temporally ordered multimodal units, \emph{OmniPick} selects the logically retained history, and \emph{OmniPage} maintains the physical KV layout used by the transformer loop.}
  \Description{A system overview showing the omni frontend creating multimodal units, the transformer and KV cache read-update loop, OmniPick updating logical retention, OmniPage managing physical KV placement, and text or optional audio output.}
  \label{fig:design-overview}
\end{figure}

\begin{figure*}[t]
  \centering
  \includegraphics[width=\textwidth]{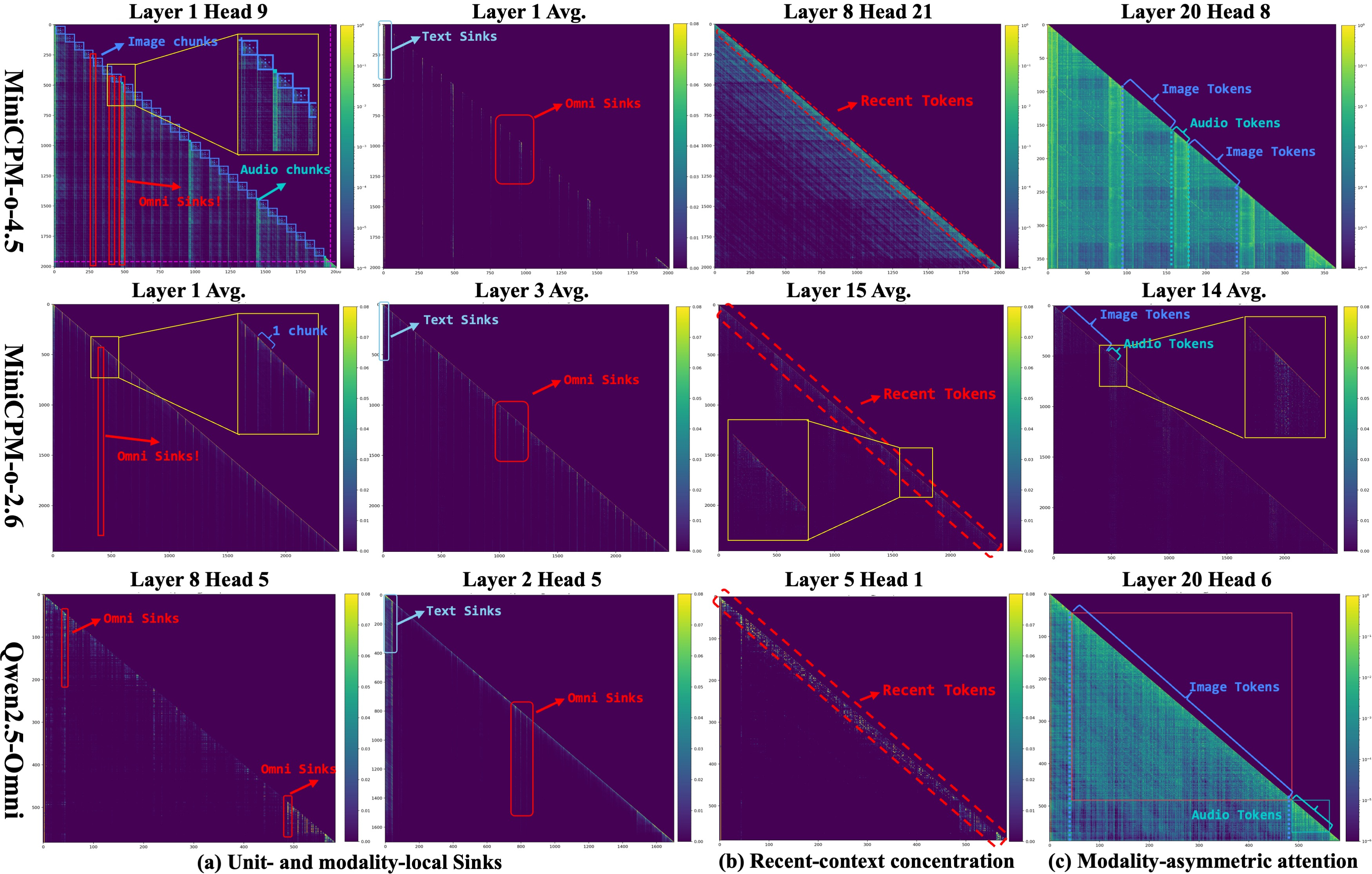}
  \caption{Selected attention traces from MiniCPM-o-4.5, MiniCPM-o-2.6, and Qwen2.5-Omni (top to bottom). }
  \Description{Three rows of attention heatmaps for MiniCPM-o-4.5, MiniCPM-o-2.6, and Qwen2.5-Omni, highlighting local sinks, recent-context concentration, and modality-asymmetric attention. }
  \label{fig:sink-traces}
\end{figure*}

\paragraph{Workflow.}
Figure~\ref{fig:design-overview} illustrates the runtime workflow. The \emph{Omni Frontend} tokenizes text, encodes visual and audio inputs, and assembles their representations into temporally ordered units with modality and structural spans. The transformer backbone processes these units through a persistent KV read--update loop during prefill and decoding~\textcircled{1}. When the accumulated history crosses the configured retention threshold, \emph{OmniPick} (\S\ref{sec:omnipick}) updates the logical cache~\textcircled{2}: it preserves complete recent units, the protected system prefix, and selected structural and modality-local sink spans from older units, then removes the remaining ranges and reindexes surviving positions. \emph{OmniPage} (\S\ref{sec:omnipage}) manages the physical storage backing KV reads and writes~\textcircled{3}. It uses retention metadata to guide placement and plans selective relocation after logical eviction. OmniPage moves retained KV entries only if the plan reduces active page or subblock counts, or meets the reduction threshold. This changes the physical layout while preserving the retained history and its logical positions for subsequent inference. The model produces text responses and may additionally generate speech through its optional audio-output path.

\endgroup

\section{OmniPick design}
\label{sec:omnipick}
OmniPick retains complete recent units and selected spans from
older units (\S\ref{subsec:retention-planning}). After removing
unselected tokens, it updates the positions of retained tokens and their metadata to continue inference with the shortened history (\S\ref{subsec:logical-deletion}).

\paragraph{Key insight: structured attention in omni-modal streams.} 
\label{subsec:structured-attention}
To understand which parts of the history should be retained, we run MiniCPM-o-4.5, MiniCPM-o-2.6, and Qwen2.5-Omni on 10 LongVALE videos~\cite{geng2025longvale} using Hugging Face Transformers~\cite{wolf2020transformers} on an NVIDIA RTX 4090 GPU with CUDA 13.2, collecting attention weights from the LLM backbone and averaging the results over the 10 videos for each model. Figure~\ref{fig:sink-traces} presents representative examples from selected layer-head pairs, alongside layer-level averages where indicated. These examples highlight three typical, recurring attention patterns shared across the evaluated models. Although their strength varies across layers and heads, these patterns provide a common basis for OmniPick's retention design.

\paragraph{Unit- and modality-local sinks.} Attention sinks can occur at both initial and internal tokens in language models~\cite{zhang2025catchtag}. Figure~\ref{fig:sink-traces}(a) shows narrow vertical bands near unit starts and modality boundaries, indicating that tokens at these locations attract attention from later units. These sinks recur within the history rather than appearing only at the initial text prefix. We refer to them as \emph{omni sinks}. We hypothesize that some of these tokens serve as local aggregation points for information within a unit, allowing later queries to access earlier context. Their repeated occurrence near structural boundaries motivates retaining small spans at these locations when evicting older content. OmniPick uses unit and modality boundaries to locate these candidate sink spans without collecting attention scores during inference. 

\paragraph{Recent-context concentration.} Figure~\ref{fig:sink-traces}(b) shows strong attention near the causal diagonal, indicating that queries frequently attend to nearby preceding tokens. This observation motivates retaining recent context alongside older sink spans. OmniPick keeps selected recent units in full so that image and audio tokens from the same unit remain available together. Using unit boundaries also avoids cutting through a chunk at an arbitrary token position.

\paragraph{Modality-asymmetric attention.} Figure~\ref{fig:sink-traces}(c) shows distinct attention intensities across image and audio spans. For example, the displayed MiniCPM-o-4.5 head assigns higher attention weights to the marked audio regions than to neighboring image regions. These differences suggest that retention should account for modality roles within a unit, motivating modality-specific budgets rather than an identical retention rule for every modality. OmniPick therefore uses separate retention budgets for image, audio, and text spans.

Together, these observations motivate a unit- and modality-aware retention policy combining local sinks, recent units, and modality-specific budgets. The heatmaps provide qualitative design evidence; we next describe how OmniPick combines these choices under a limited history budget.

\begin{figure}[t]
  \centering
  \includegraphics[width=\columnwidth]{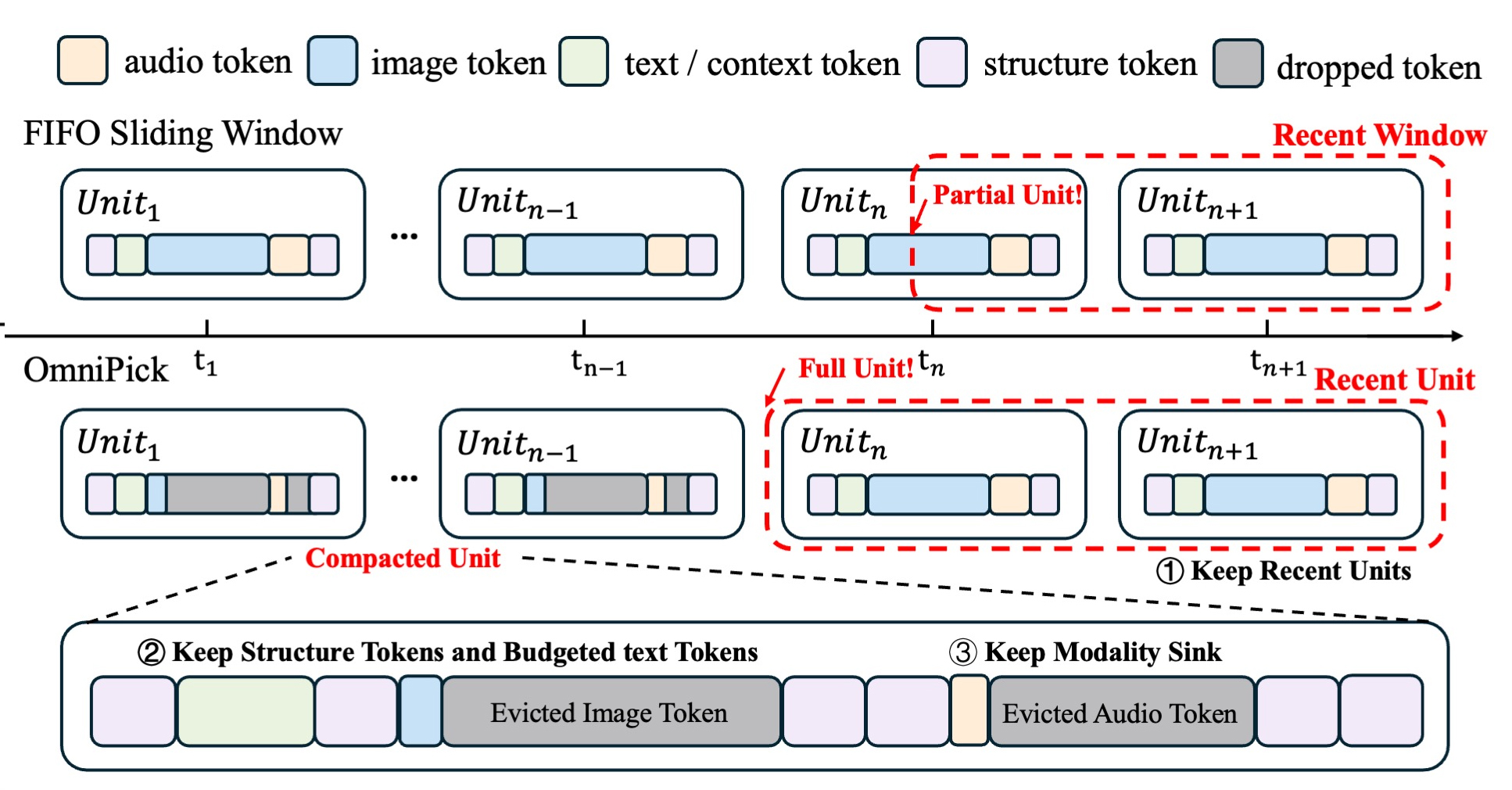}
  \caption{OmniPick retains a recent window of complete units and selected sink spans from older units. FIFO in this schematic denotes token-level sliding.}
  \Description{A schematic of OmniPick retaining recent complete units and selected partial spans, with FIFO shown as a positional comparison.}
  \label{fig:omnipick}
\end{figure}

\subsection{Structure-aware retention planning}
\label{subsec:retention-planning}
OmniPick starts retention when the history exceeds the high watermark and uses the low watermark as its trimming target. Figure~\ref{fig:omnipick} illustrates the main idea: retain recent units in full and keep only selected spans from older units. The system prefix remains protected throughout.

\paragraph{Keeping recent units intact.} OmniPick retains the most recent units in full~\textcircled{1}, while token-level sliding retains a fixed number of recent tokens regardless of unit boundaries. As shown in Figure~\ref{fig:omnipick}, the token-level window cuts through the image span of $u_n$, retaining only part of the unit. OmniPick instead keeps both $u_n$ and $u_{n+1}$ intact, preserving their image, audio, text, and boundary tokens together.

\paragraph{Selecting spans from older units.} For units not retained in full, OmniPick visits them from oldest to newest. Consider $u_{n-1}$ in Figure~\ref{fig:omnipick}: OmniPick retains its boundary tokens~\textcircled{2} and short prefixes of its image and audio spans as candidate sinks~\textcircled{3}, then discards the remaining modality tokens. To account for differences in modality importance, it uses separate prefix lengths for image and audio spans. Text and previous model responses are retained from oldest to newest until their shared token budget is exhausted. For models that represent an image as a source image and local slices, we retain the source-image span in full and keep only the sink prefix of each slice.

If the selected history still exceeds the low watermark,
OmniPick drops units, oldest first, prioritizing partially
retained units over complete ones. It stops when the target
is met or only one unit remains. The low watermark is
therefore a trimming target rather than a strict upper bound.

OmniPick generates a plan that lists the spans to retain and the token ranges to delete, without collecting attention scores during inference.

\subsection{Applying the retention plan}
\label{subsec:logical-deletion}
Removing tokens from within a unit changes both token
positions and the boundaries of its remaining spans.
OmniPick therefore updates the cache and the unit records
together. It deletes the unselected ranges and shifts each
retained token's logical position by the number of deleted
tokens before it. The runtime applies the corresponding
RoPE adjustment to cached keys.
For each partially retained unit, OmniPick recomputes its
length and span offsets while preserving the spans' modality
and boundary labels. This allows the next retention round
to identify image, audio, text, and boundary spans in the
shortened unit. These updates leave surviving KV entries
in their physical slots; OmniPage separately decides
whether to relocate them (\S\ref{sec:omnipage}).

\begin{figure}[t]
  \centering
  \includegraphics[width=\columnwidth]{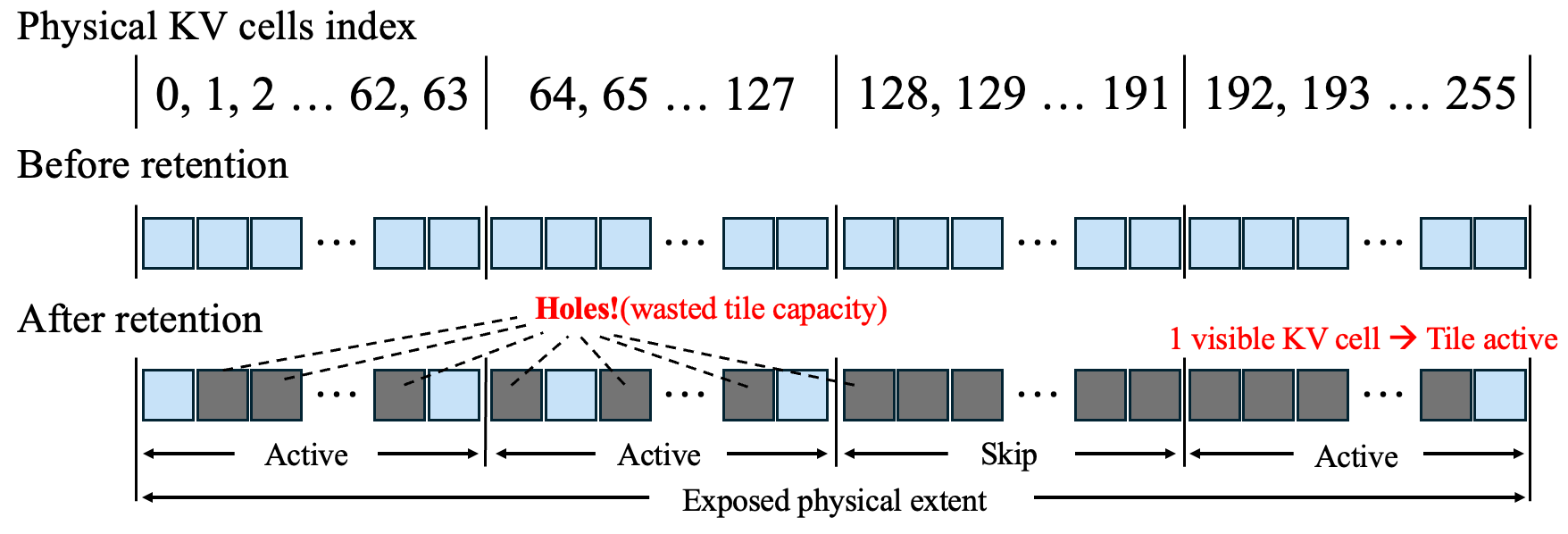}
  \caption{Logical eviction leaves masked holes between surviving KV entries. Sparse holes keep the exposed physical extent large.}
  \Description{Physical KV cells before and after retention, showing masked holes and active or skipped tiles within the exposed physical extent.}
  \label{fig:omnipage-problem}
\end{figure}

\section{OmniPage: physical KV management}
\label{sec:omnipage}
\begingroup
\emergencystretch=1em

\paragraph{Translating logical retention into physical savings.}
\label{subsec:layout-challenge}
OmniPick reduces retained tokens, but the
remaining KV entries can still be scattered across the cache.
OmniPage addresses two problems to reduce attention cost.

\paragraph{$\bullet$ Fragmentation after logical eviction.} As shown in Figure~\ref{fig:omnipage-problem}, deleting tokens leaves holes between retained KV entries without moving them. Attention kernels process KV entries in groups called \emph{tiles}. On paths that support tile skipping, a tile can be skipped only when all its entries are masked. Even one retained sink token can therefore keep a tile active. Scattered entries can also keep the physical span---the range of KV slots presented to attention---large. Consequently, retaining fewer tokens does not necessarily reduce attention work by the same proportion.

\begin{figure}[t]
  \centering
  \includegraphics[width=\columnwidth]{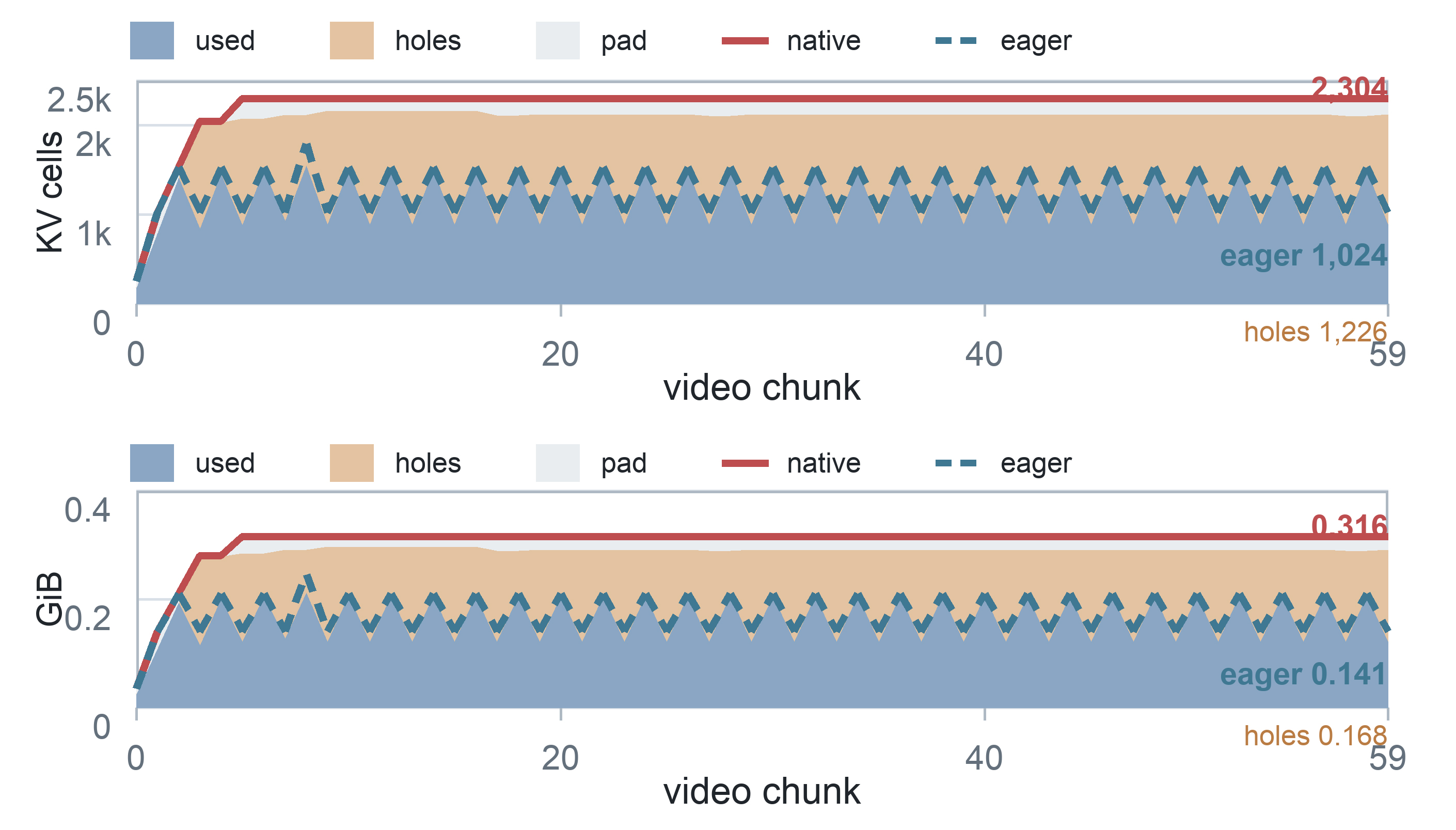}
  \caption{Native and eager layouts under OmniPick retention in a 60-chunk MiniCPM-o-4.5 Metal pressure run.}
  \Description{Physical extent, occupied cells, and holes for native and eager layouts, alongside their exposed and useful byte footprints.}
  \label{fig:physical-diagnostics}
\end{figure}

\paragraph{$\bullet$ Compaction-movement trade-off.}
Compaction can recover the space left by logical eviction. A straightforward approach is \emph{eager compaction}, which packs all retained KV entries into a contiguous prefix after every successful eviction. Figure~\ref{fig:physical-diagnostics} compares this approach with native deletion under the same OmniPick retention policy in a 60-chunk MiniCPM-o-4.5 run on an Apple M2 Pro with Metal. Eager compaction reduces the final physical span from 2,304 to 1,024 KV slots---a $2.25\times$ difference. But it requires K/V copies and synchronization after each eviction. This motivates selective migration: reducing fragmentation while moving only part of the retained history.

\paragraph{Key insight.}
OmniPick's retention decisions provide a signal for organizing physical KV storage: newly arriving content and retained spans from older units need not share the same pages. OmniPage therefore groups KV entries by retention state. Pages align with attention tiles, so packing retained entries into fewer tiles can leave fully empty tiles that supported kernels can skip. Selective migration reduces fragmentation without repacking the entire cache.

OmniPage combines \emph{retention-aware, tile-aligned paging} (\S\ref{subsec:retention-aware-paging}) to guide KV placement with \emph{selective KV migration} (\S\ref{subsec:selective-migration}) to pack scattered entries into fewer tiles or a shorter physical span. Both mechanisms preserve OmniPick's selected history and the logical positions of retained entries.

\begin{figure*}[t]
  \centering
  \includegraphics[width=0.92\textwidth]{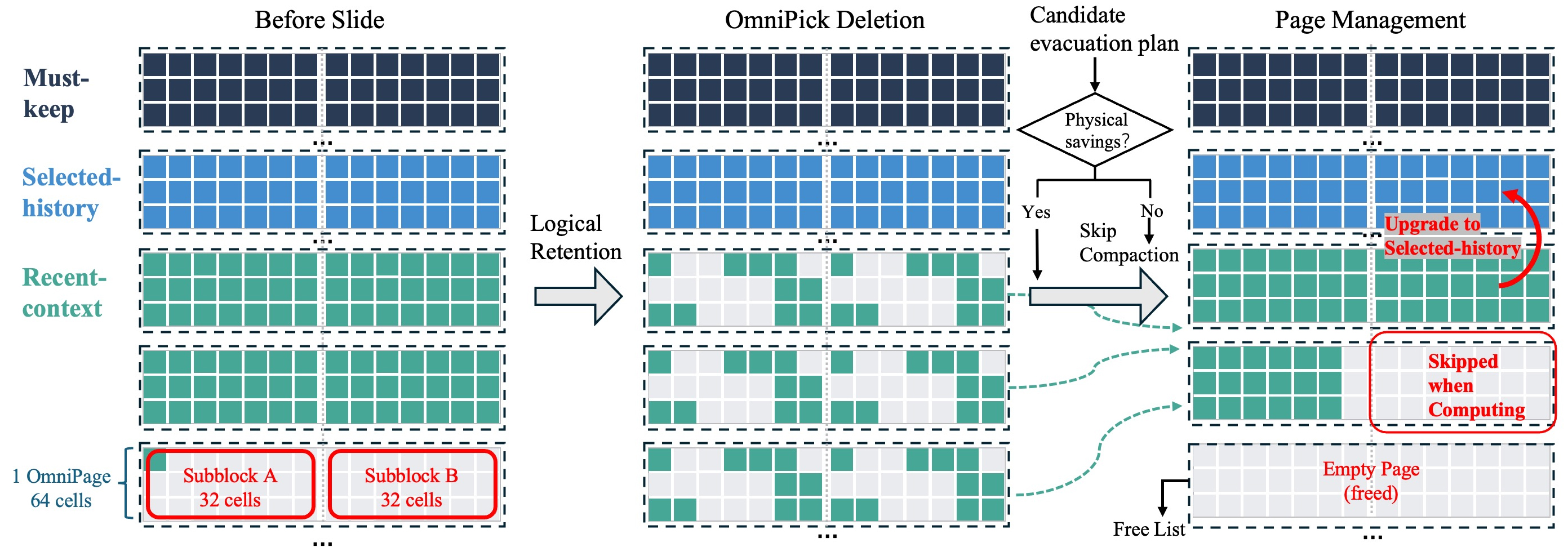}
  \caption{Class-aware OmniPage placement and selective migration, illustrated with 64-cell pages and 32-cell subblocks.}
  \Description{Sparse KV cells organized by allocation class, with retained cells selectively relocated to reclaim pages within the existing tensor.}
  \label{fig:layout}
\end{figure*}

\subsection{Retention-aware, tile-aligned paging}
\label{subsec:retention-aware-paging}

\paragraph{Organizing KV into tile-aligned pages.}
OmniPage partitions physical KV storage into fixed-size, contiguous groups of slots called \emph{pages}. OmniPick groups tokens by input chunks or responses, while OmniPage groups their KV entries by physical location. Page size is tied to the backend's KV-axis tile width, so that empty pages can be skipped by kernels with tile-skipping support. For kernels with smaller tiles, it also tracks which subblocks within each page still contain KV entries.

For example, in our Metal experiments, Figure~\ref{fig:layout} illustrates a 64-slot page containing two 32-slot subblocks. This organization corresponds to one 64-slot KV tile in the inspected Metal tiled path, or two 32-slot KV tiles in its vector path. Page and subblock sizes follow the attention kernels used by each backend. An empty page can therefore be skipped at either granularity. A page with one empty subblock provides a finer-grained skipping opportunity on the vector path, whereas the 64-slot path still requires the entire page to be empty before skipping its tile. The tile-oriented execution model follows the same distinction between logical visibility and kernel work exposed by FlashAttention-style implementations~\cite{dao2022flashattention,ye2025flashinfer}.

On Metal paths with tile-skipping support, packing retained entries into fewer tiles allows empty tiles to be skipped. For selected CUDA configurations, migration instead targets a shorter physical span, reducing the range of KV slots traversed by attention. Thus, the same page-based organization supports different backends, while the migration criterion reflects each kernel's cache access.

\paragraph{Placing entries according to retention state.} Tile alignment alone does not prevent fragmentation. If newly arriving content shares pages with spans retained from older units, later eviction can leave a few retained entries scattered across many pages. OmniPage uses the three retention groups in Figure~\ref{fig:layout} to guide physical placement: \emph{must-keep} content includes the protected system prefix; \emph{selected-history} contains spans retained from older units; and \emph{recent-context} contains recent units that are still retained in full.

New entries enter recent-context. After retention, OmniPage
packs the surviving entries into fewer pages and promotes
fully packed pages to selected-history. Partially filled
pages remain in recent-context, while empty pages can be
reused for new entries. These moves are performed only
when the migration plan meets the layout criteria
(\S\ref{subsec:selective-migration}).

\subsection{Selective and efficient KV migration}
\label{subsec:selective-migration}

After eviction, retained KV entries may occupy only a few slots in each page. OmniPage first plans which entries to move and where to place them. It then checks whether the proposed moves reduce the number of active pages or subblocks, or shorten the physical span enough to meet the configured threshold. Only accepted plans trigger K/V copies.

\paragraph{Packing scattered entries.}
OmniPage targets entries retained from partially evicted units, leaving pages filled with entries in place. It scans the candidate entries in physical order and assigns them to earlier empty slots. A destination page must be empty or contain only entries of the same allocation class. Filling these slots can leave later pages or subblocks empty, while entries outside the plan remain in place.

\paragraph{Checking layout improvement.}
Filling holes doesn't always reduce attention work: moving entries between two pages may still leave both pages active. OmniPage therefore simulates the planned moves and counts the active pages and subblocks before and after migration. A region is active if it contains any retained entry. The plan is accepted only if either count decreases. With tile-aligned pages, empty regions can be skipped by kernels with tile-skipping support.

Some kernels traverse the physical span, so emptying interior pages may not reduce the range they process. For the selected CUDA configurations, OmniPage instead uses an \emph{ordered-suffix} mode. It scans backward from the end of the cache to find a suffix whose packing would shorten the page-rounded physical span by at least a configured number of pages. It packs only the retained entries in that suffix, preserving their order and leaving the preceding entries untouched. If no suffix meets the threshold, no migration occurs. This mode replaces class-based placement and can move entries from any retention group.

\paragraph{Executing accepted moves efficiently.}
For an accepted plan, OmniPage only copies selected KV entries. The plan specifies a source-destination mapping shared across the affected layers' KV tensors. On CUDA, the runtime gathers source rows into scratch space before writing them to destinations. This prevents a write from overwriting a source row needed by another move. Other supported backends apply the mapping through their tensor-copy primitives.

Migration runs after logical deletion and position updates, before inference resumes. The runtime synchronizes pending work, performs the copies, and updates storage metadata. Each moved entry retains its logical position, pending position shift, sequence membership, and allocation class. Migration therefore changes where KV entries are stored without changing the history selected by OmniPick.

\endgroup

\section{Implementation and evaluation}
\label{sec:evaluation}
\begingroup
\emergencystretch=2em
We implement OmniTide in llama.cpp-omni~\cite{llamacppomni2026}, adding approximately 9.6K code lines across the runtime, model runners, and backend extensions, excluding experiment scripts and tests. Model-specific input adapters translate each model’s omni-modal input structure into a shared unit/span representation, allowing Qwen and MiniCPM to reuse the implementation. We extend llama.cpp's KV memory interface with allocation classes, range promotion, and selective migration, so model runners can invoke physical maintenance without implementing backend-specific KV movement. OmniTide supports Qwen2.5-Omni-3B/7B~\cite{xu2025qwen25omni} and MiniCPM-o-4.5~\cite{cui2026minicpmo45} on CUDA and Metal.

We evaluate latency, quality, memory and KV storage, examining the quality-efficiency tradeoff and the contributions of retention budgets and physical maintenance.

\subsection{Experimental setup}
\label{subsec:eval-setup}

\paragraph{Hardware and precision.}
We evaluate CUDA on an NVIDIA RTX 4090 (24\,GB) and Metal on an Apple M2 Pro (32\,GB unified memory). We use Q4\_K\_M backbone weights, Q8\_0 Qwen projectors, FP16 MiniCPM vision/audio projectors, and FP16 KV storage. FlashAttention is enabled for the main timing comparisons.

\paragraph{Workloads.}
We evaluate timestamped multiple-choice QA on StreamingBench~\cite{lin2025streamingbench}, multi-turn video dialogue on SVBench~\cite{yang2025svbench}, and sports commentary on LiveSports-3K-CC~\cite{chen2025livecc}. Main comparisons use inputs within each model's training context length. Video is sampled at 1 FPS with aligned one-second audio segments when applicable; history persists across inputs and responses.

\begin{figure*}[t]
  \centering
  \includegraphics[width=\textwidth]{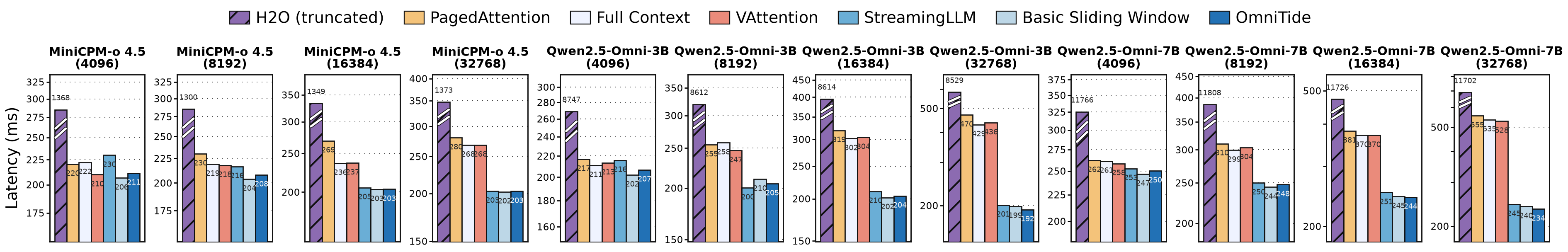}\par
  {\small\textbf{(a)} Normal streaming-update latency}\par\vspace{5pt}
  \includegraphics[width=\textwidth]{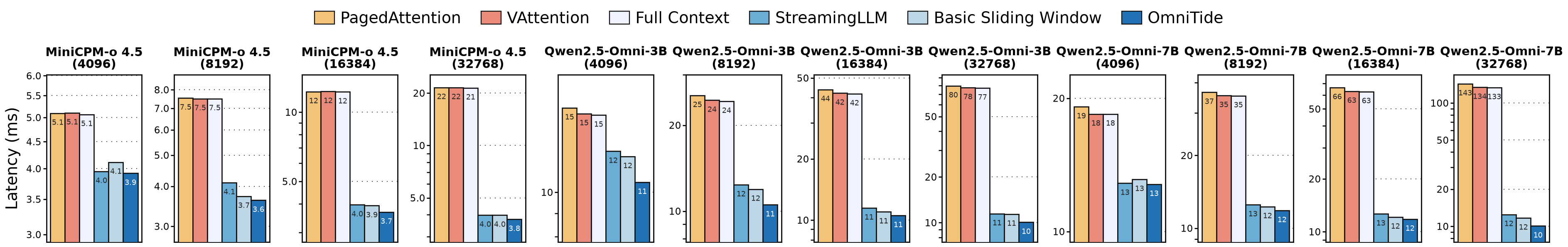}\par
  {\small\textbf{(b)} Separately profiled attention-kernel latency}
  \caption{CUDA latency at 4K--32K no-slide checkpoints. Panels show three models with four checkpoints each.}
  \Description{CUDA streaming-update and attention-kernel latency for MiniCPM-o-4.5 and Qwen2.5-Omni-3B/7B at four history lengths.}
  \label{fig:cuda-latency}
\end{figure*}

\paragraph{Baselines and configurations.}
(i) No-slide (\emph{Full Context}) retains the complete history.
(ii) Unit-level FIFO (\texttt{basic}) evicts the oldest complete units.
(iii) Token-level sliding (\texttt{token-basic}) is a strict sliding-window control that evicts the oldest tokens regardless of unit or span boundaries.
(iv) StreamingLLM~\cite{xiao2024streamingllm} retains a positional sink prefix and a recent suffix.
(v) H$_2$O~\cite{zhang2023h2o} provides an attention-based retention reference.
(vi) PagedAttention~\cite{kwon2023vllm} and
(vii) vAttention~\cite{prabhu2025vattention} provide physical-layout references.

\emph{OmniPick only} uses native KV layout with OmniPage disabled; \emph{OmniTide} enables both components. In the physical ablation (\S\ref{subsec:eval-page-ablation}), eager compaction serves as an \emph{attention-time oracle}: it repacks all retained KV entries into a contiguous layout while its repacking cost is excluded. Our default high/low watermarks are 2500/1800 tokens; StreamingLLM uses a two-token sink prefix.

\paragraph{Metrics.}
We report latency per streaming update, mean stream-loop time per session, attention-kernel latency, and KV storage (retained cells and physical span). Task quality is measured by multiple-choice accuracy on StreamingBench and judge scores on SVBench and LiveSports, both reported on a 0--100 scale. We profile kernels separately and disable profiling when measuring streaming-update latency to avoid instrumentation overhead.

\begin{figure*}[t]
  \centering
  \includegraphics[width=\textwidth]{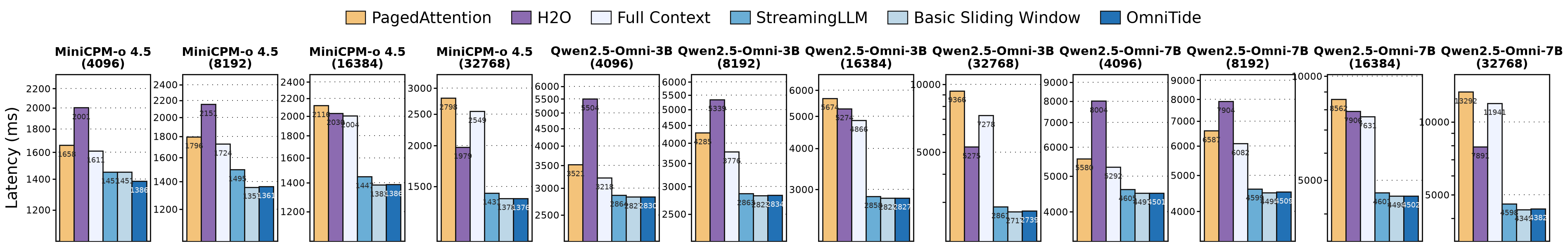}\par
  {\small\textbf{(a)} Normal streaming-update latency}\par\vspace{5pt}
  \includegraphics[width=\textwidth]{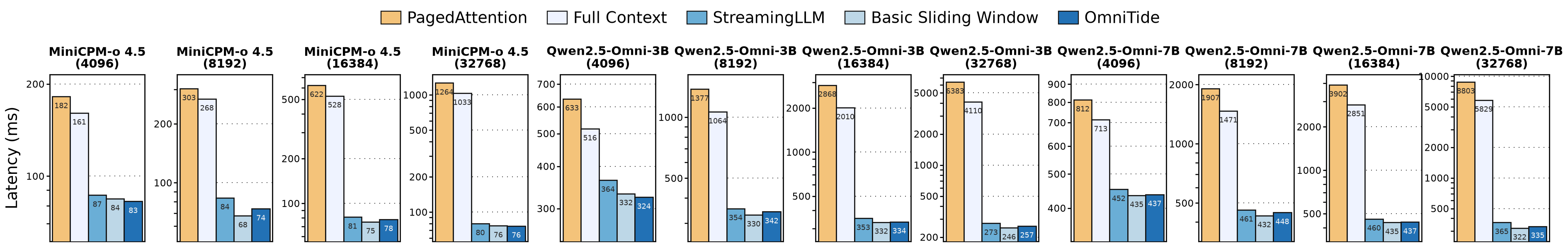}\par
  {\small\textbf{(b)} Separately profiled attention-kernel latency}
  \caption{Metal latency at matched no-slide checkpoints, with normal update timing above separate attention profiling.}
  \Description{Metal streaming-update and attention-kernel latency for three models at 4K, 8K, 16K, and 32K no-slide histories.}
  \label{fig:metal-latency}
\end{figure*}

\begin{table}[t]
\centering
\caption{Task-quality summary. Each cell reports \textbf{StreamingBench / SVBench / LiveSports-3K-cc}. StreamingBench reports percentages; SVBench reports 0--100 Terra dialogue scores; LiveSports reports 0--100 Terra semantic-alignment scores.}
\label{tab:quality}
\small
\setlength{\tabcolsep}{3pt}
\begin{tabular}{@{}>{\centering\arraybackslash}p{0.22\columnwidth}*{3}{>{\footnotesize\centering\arraybackslash}p{0.24\columnwidth}}@{}}
\toprule
 & \multicolumn{2}{c}{\small\textbf{Qwen2.5-Omni}}
 & {\small\textbf{MCPM-o-4.5}} \\
\cmidrule(lr){2-3}\cmidrule(l){4-4}
\textbf{Method}
 & {\small\textbf{3B}}
 & {\small\textbf{7B}}
 & {\small\textbf{9B}} \\
\midrule
No-slide & 62.3 / 43.4 / 30.2 & 52.0 / \textbf{55.1} / 31.2 & 74.1 / 43.4 / 17.7 \\
FIFO / basic & 58.1 / 42.9 / 29.2 & 49.8 / 53.9 / 30.2 & 69.0 / 40.6 / 17.3 \\
Token-basic & 56.2 / 42.2 / 28.9 & 47.8 / 52.6 / 30.3 & 57.0 / 40.1 / 16.8 \\
StreamingLLM & 59.6 / 42.6 / 28.2 & 50.5 / 47.8 / 31.4 & 70.1 / 41.3 / 17.5 \\
H2O & 55.8 / 40.1 / 28.2 & 47.0 / 49.2 / 29.2 & 52.7 / 35.1 / 17.0 \\
\addlinespace[2pt]
\textbf{Ours} & \textbf{62.8} / \textbf{43.5} / \textbf{30.2} & \textbf{52.4} / 54.8 / \textbf{33.3} & \textbf{75.0} / \textbf{45.1} / \textbf{18.8} \\
\bottomrule
\end{tabular}
\end{table}

\subsection{Overall performance}
\label{subsec:eval-overall}

\subsubsection{Latency}
\label{subsec:eval-efficiency}
We evaluate how per-update latency changes as no-slide history grows from 4K to 32K tokens (Figures~\ref{fig:cuda-latency} and~\ref{fig:metal-latency}). Upper panels report normal streaming updates; lower panels report separately profiled FlashAttention kernels. H2O requires attention scores that the FlashAttention kernels used here do not expose. It is therefore evaluated without FlashAttention, contributing to its higher update latency, and omitted from the kernel panels. vAttention relies on CUDA virtual memory management~\cite{prabhu2025vattention} and is therefore evaluated only on CUDA. \textbf{OmniTide keeps update and attention latency comparatively stable as history grows, reducing long-context latency on both backends.}

\paragraph{CUDA.}
At 32K, OmniTide reduces update latency from approximately 429 to 192\,ms for Qwen-3B, 535 to 234\,ms for Qwen-7B, and 268 to 203\,ms for MiniCPM. Qwen-7B attention falls from approximately 133 to 10\,ms. OmniPick reduces the retained history, while OmniPage compacts its physical layout to reduce attention work. Streaming-update speedups are smaller than kernel speedups because each update also includes computation outside attention. We examine OmniPage's contribution separately in \S\ref{subsec:eval-page-ablation}.

\paragraph{Metal.}
At 32K, Qwen-3B update latency decreases from approximately 7.3 to 2.7\,s and Qwen-7B from 11.9 to 4.3\,s. Across the checkpoints, MiniCPM full-context attention grows from approximately 161 to 1,033\,ms, whereas OmniTide remains at 74--83\,ms. The similar scaling trend on both backends supports bounding retained history as the main source of long-context savings. Other bounded-history policies achieve similar latency, making retention quality a key differentiator, as evaluated next.

\begin{figure}[t]
  \centering
  \includegraphics[width=\columnwidth]{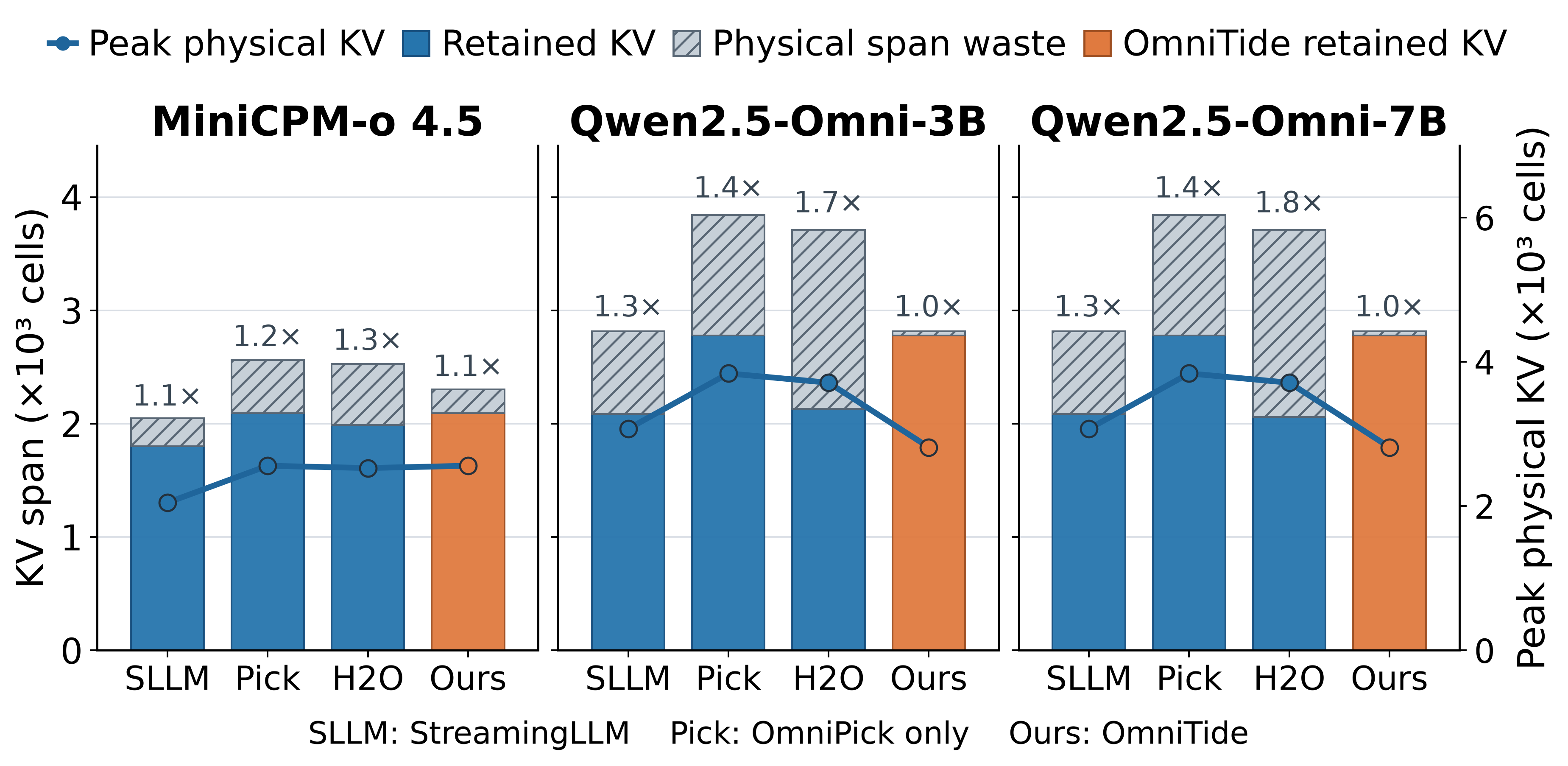}
  \caption{CUDA KV layout at the 32K checkpoint. Stacked bars show useful cells and physical-span waste (left axis); the line shows cumulative peak physical span (right axis). Labels above bars give span/useful-cell ratios.}
  \Description{Three model panels comparing useful KV, physical span, and peak span for StreamingLLM, OmniPick only, H2O, and OmniTide.}
  \label{fig:physical-memory}
\end{figure}

\begin{figure*}[t]
  \centering
  \includegraphics[width=\textwidth]{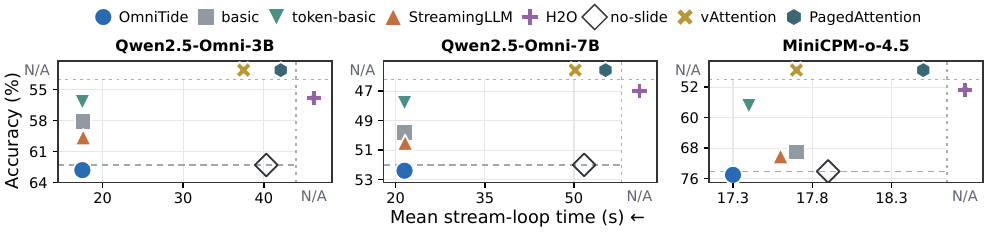}
  \caption{StreamingBench quality--efficiency tradeoff. Lower left is better: session time decreases to the left and accuracy increases downward. Dashed horizontal lines mark no-slide accuracy; N/A bands denote unreported metrics.}
  \Description{Three panels for Qwen2.5-Omni-3B, Qwen2.5-Omni-7B, and MiniCPM-o-4.5 comparing mean stream-loop time and accuracy. H2O appears in the latency N/A band; PagedAttention and vAttention appear in the accuracy N/A band. OmniTide combines low session time with high accuracy among the plotted configurations.}
  \label{fig:quality-efficiency}
\end{figure*}

\subsubsection{Task quality}
\label{subsec:eval-quality}
We compare retention quality on StreamingBench, SVBench, and LiveSports in Table~\ref{tab:quality}. \textbf{OmniPick outperforms FIFO and token-level sliding across all nine model--benchmark settings.}

\paragraph{Comparison with bounded-history policies.}
On StreamingBench, unit-level FIFO improves over token-level sliding by 1.9, 2.0, and 12.0 percentage points for Qwen-3B, Qwen-7B, and MiniCPM, respectively. These gains highlight the importance of preserving complete multimodal units during eviction. OmniPick further improves over FIFO by 4.7, 2.6, and 6.0 points: beyond preserving complete recent units, its selected boundary, sink, and modality spans retain older context that FIFO discards. These results evaluate the combined policy rather than individual span types.

OmniPick remains close to no-slide: StreamingBench differences are +0.5, +0.4, and +0.9 points, while SVBench and LiveSports differences range from $-0.3$ to $+2.1$ points. We report these differences descriptively rather than claim an accuracy advantage over full-context retention.

\begin{table}[t]
\centering
\caption{End-to-end session latency on CUDA, excluding model initialization and system-prompt prefill. }
\label{tab:latency}
\small
\setlength{\tabcolsep}{4pt}
\begin{tabular}{@{}lrrr@{}}
\toprule
Model & \makecell{No-slide\\(s)} & \makecell{OmniTide\\(s)} & Speedup \\
\midrule
Qwen2.5-Omni-3B & 40.33 & 17.61 & 2.290$\times$ \\
Qwen2.5-Omni-7B & 51.71 & 21.52 & 2.403$\times$ \\
\bottomrule
\end{tabular}
\end{table}

\subsubsection{Physical KV storage}
\label{subsec:eval-memory}
We evaluate KV storage in Figure~\ref{fig:physical-memory}, separating useful retained cells from the physical span exposed to attention. \textbf{OmniPage reduces physical span by 26.7\% for Qwen and 10.0\% for MiniCPM while preserving the retained-cell count.}

Both Qwen models retain 2,778 useful cells, while OmniPage reduces physical span from 3,840 to 2,816 cells. MiniCPM retains 2,095 cells within a span of 2,304 rather than 2,560 cells. Qwen's cumulative peak span also decreases; MiniCPM's is unchanged. OmniPage fills holes left by logical deletion through selective relocation, reducing physical span without changing retained history.

\subsubsection{End-to-end latency}
\label{subsec:eval-session-speedup}
We measure CUDA session latency over 355 sessions (Table~\ref{tab:latency}) excluding model initialization and system-prompt prefill. \textbf{OmniTide achieves 2.29--2.40$\times$ speedup over no-slide}, reducing mean latency from 40.33 to 17.61\,s for Qwen-3B and from 51.71 to 21.52\,s for Qwen-7B. Bounded history and a consolidated KV layout reduce attention work across successive updates.

\begin{figure}[t]
  \centering
  \includegraphics[width=\columnwidth]{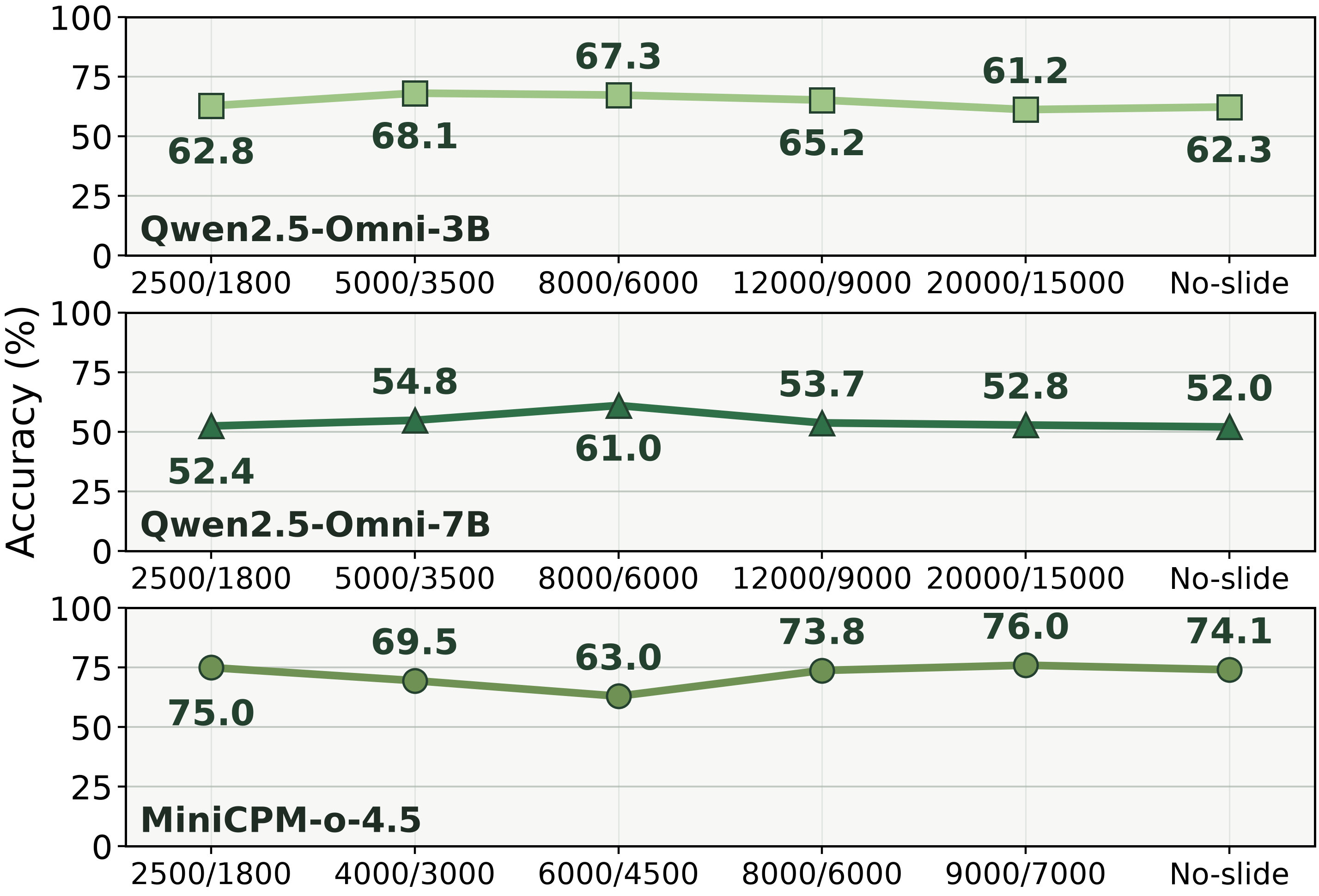}
  \caption{Watermark sensitivity on StreamingBench. The vertical axis is accuracy (\%); horizontal labels give high/low watermarks, with no-slide as a reference.}
  \Description{Three accuracy curves across high/low watermark settings, with no-slide references.}
  \label{fig:retention}
\end{figure}

\subsection{Quality--efficiency tradeoff}
\label{subsec:eval-tradeoff}
We jointly evaluate StreamingBench accuracy and mean complete-session stream-loop time in Figure~\ref{fig:quality-efficiency}. Timing includes input processing, prefill, answer generation, and KV maintenance, but excludes initialization and system-prompt prefill. Accuracy and timing use the same runs; \textbf{OmniTide preserves more task quality than sliding-window baselines at comparable session cost.}

For Qwen-3B and Qwen-7B, mean time decreases from 40.3 to 17.5\,s and from 51.7 to 21.5\,s relative to no-slide, while accuracy rises from 62.3\% to 62.8\% and from 52.0\% to 52.4\%. For MiniCPM, time decreases only from 17.9 to 17.3\,s, while accuracy rises from 74.1\% to 75.0\%. Basic, token-basic, and StreamingLLM have similar costs but lower accuracy. Their similar costs are consistent with bounded history limiting execution work, while OmniPick's structured selection preserves more useful context at that cost. The smaller MiniCPM timing gain limits how broadly Qwen's speedups can be generalized.

N/A marks H2O latency because it runs without FlashAttention, and PagedAttention/vAttention accuracy because they change physical layout without selecting history.

\begin{figure}[t]
  \centering
  \includegraphics[width=\columnwidth]{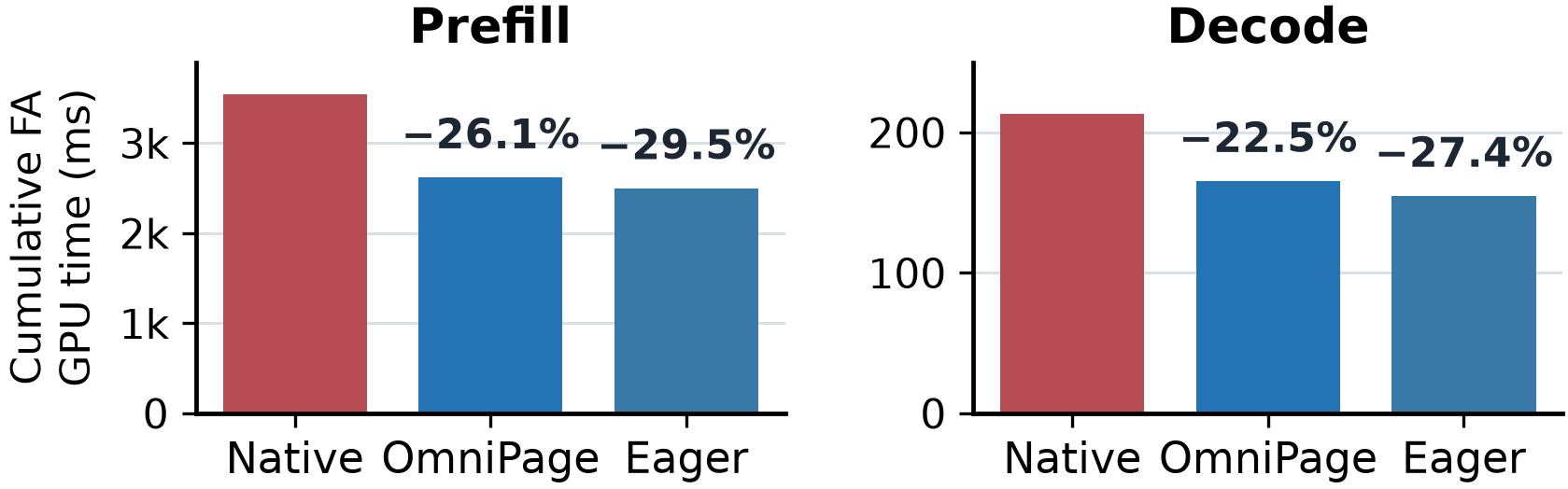}
  \caption{Cumulative attention GPU time under identical retention. OmniPage includes migration and synchronization costs; Eager excludes them.}
  \Description{Prefill and decode cumulative FlashAttention GPU times, with OmniPage reductions of 26.1 and 22.5 percent relative to native layout.}
  \label{fig:omnipage-ablation}
\end{figure}

\subsection{Ablation and sensitivity}
\label{subsec:eval-ablation}

\subsubsection{Watermark sensitivity}
\label{subsec:eval-watermark}
We evaluate OmniPick's sensitivity to high/low watermarks on StreamingBench (Figure~\ref{fig:retention}). \textbf{Larger retention budgets do not monotonically improve accuracy; the best tested setting depends on the model.}

For Qwen-3B, accuracy rises from 62.8\% at the default 2500/1800 setting to 68.1\% at 5000/3500, then falls to 61.2\% at 20000/15000. Qwen-7B peaks at 61.0\% with 8000/6000, compared with 52.4\% at the default and 52.8\% at 20000/15000. MiniCPM also varies non-monotonically: accuracy falls from 75.0\% at 2500/1800 to 63.0\% at 6000/4500 before reaching 76.0\% at 9000/7000. These results show that retention quality depends on the watermark configuration, motivating model-specific tuning rather than simply increasing the retained history.

\subsubsection{Effectiveness of OmniPage}
\label{subsec:eval-page-ablation}
We compare cumulative prefill and decode attention GPU time under native layout, OmniPage, and eager compaction (Figure~\ref{fig:omnipage-ablation}). \textbf{OmniPage reduces prefill and decode attention time by 26.1\% and 22.5\% relative to native layout.}

Eager compaction reduces prefill and decode attention time by 29.5\% and 27.4\% respectively, providing an oracle that fully packs retained KV entries without repacking cost. OmniPage achieves 26.1\% and 22.5\% reductions, capturing 88.5\% and 82.1\% of the oracle's latency reduction through selective migration. These results indicate OmniPage recovers most of the attention benefit of full compaction.

\par
\endgroup

\section{Related work}
\begingroup
\emergencystretch=2em
\paragraph{Multimodal KV retention and compression.}
Multimodal KV compression exploits modality-dependent relevance, redundancy, and layer-wise importance~\cite{wan2024lookm,wan2025meda,huang2025aircache,wang2025prefixkv}, or frequency-domain outliers~\cite{yang2026flashcache}. Video-oriented methods compress and retrieve historical visual KV states~\cite{xiao2026mukv,chen2026flexmem}. Visual token reduction operates before LLM processing~\cite{yang2025visionzip,wang2026stc} or prunes tokens within the LLM~\cite{chen2024fastv}; STC reuses encoder features. OmniPick instead uses structural metadata to retain complete recent audiovisual units and typed boundary and sink spans from older history.

\paragraph{KV memory management and attention execution.}
PagedAttention and vAttention optimize KV allocation~\cite{kwon2023vllm,prabhu2025vattention}, while GMLake addresses training-time allocation fragmentation~\cite{guo2024gmlake}. Attention and decoding kernels improve data reuse, parallelism, and scheduling~\cite{dao2022flashattention,dao2024flashattention2,shah2024flashattention3,hong2024flashdecoding,sanovar2025leanattention,ye2025flashinfer}. OmniPage addresses eviction-induced fragmentation within sessions, consolidating survivors to reduce the physical span exposed to attention.

\paragraph{KV quantization, offloading, and selective access.}
Quantization reduces KV precision~\cite{liu2024kivi,hooper2024kvquant,duanmu2024skvq,lin2025qserve}; offloading expands capacity across memory tiers~\cite{sheng2023flexgen,lee2024infinigen}; selective access fetches query-relevant entries or pages~\cite{ribar2024sparq,tang2024quest,yang2025lserve}. These techniques optimize KV representation, placement, or access, whereas OmniPick selects which interleaved omni-modal units and spans persist across updates.

\paragraph{KV cache reuse.}
Reuse systems avoid repeated prefill or duplicate storage for shared prefixes, prompt modules, and dialogue history~\cite{zheng2024sglang,gim2024promptcache,ye2024chunkattention,gao2024cachedattention,yu2025pensieve}. Mooncake manages reusable KV states across disaggregated serving~\cite{qin2025mooncake}; CacheBlend and Cache-Craft reuse retrieved chunks with selective recomputation~\cite{yao2025cacheblend,agarwal2025cachecraft}, and VLMCache exploits stable visual backgrounds~\cite{zhang2026vlmcache}. OmniTide manages growing history within a continuous session without requiring shared prefixes or repeated visual inputs.

\paragraph{Native sparse attention.}
Native sparse attention incorporates selective access into model architecture and training. NSA combines token compression and selection; VideoNSA adapts this design to video-language models~\cite{yuan2025nsa,song2026videonsa}. DeepSeek-V4.1-Flash combines compressed sparse attention with cross-layer KV reuse and low-precision caching~\cite{deepseek2026v41flash}. Adapting these learned mechanisms to existing omni-modal checkpoints would require architectural changes and training; OmniTide manages retained history and physical KV layout without retraining.

\par
\endgroup

\section{Conclusion}
\label{sec:conclusion}

This paper presents \texttt{OmniTide}, an efficient inference system that improves the accuracy--latency tradeoff for on-device streaming omni-modal applications. 
It integrates \textit{OmniPick} for structure-aware logical token retention and \textit{OmniPage} for selective physical KV consolidation to minimize memory fragmentation and attention overhead. 
Extensive experiments demonstrate its effectiveness in significantly reducing stream-loop latency and attention computation while preserving or exceeding full-context task accuracy. 
Ultimately, \texttt{OmniTide} unlocks real-time, infinite-context omni model serving on consumer-grade devices.

\bibliographystyle{ACM-Reference-Format}
\bibliography{references}
\end{document}